\documentclass[11pt]{article}

\usepackage[preprint]{acl}

\usepackage{times}
\usepackage{latexsym}
\usepackage[T1]{fontenc}
\usepackage[utf8]{inputenc}
\usepackage{microtype}
\usepackage{inconsolata}
\usepackage{graphicx}

\usepackage{amsmath}
\usepackage{amssymb}
\usepackage{booktabs}
\usepackage{multirow}
\usepackage{array}
\usepackage{xspace}
\usepackage{subcaption}
\usepackage{algorithm}
\usepackage{algpseudocode}
\usepackage{tikz}
\usetikzlibrary{positioning,arrows.meta,calc,fit,backgrounds,shapes.geometric,shapes.misc}
\usepackage{pgfplots}
\pgfplotsset{compat=1.18}
\usepgfplotslibrary{fillbetween}
\usepackage{listings}
\usepackage[most]{tcolorbox}
\tcbuselibrary{listings,breakable,skins}
\usepackage[capitalize,noabbrev]{cleveref}
\usepackage{xurl}

\graphicspath{{figures/}}

\DeclareUnicodeCharacter{221E}{\ensuremath{\infty}}
\DeclareRobustCommand{\methodsup}{\textsuperscript{\IfFontSeriesContextTF{bf}{\boldmath}{}∞}}
\newcommand{\method}{\texorpdfstring{Skill\methodsup}{Skill∞}\xspace}

\definecolor{skillboxbg}{HTML}{F5F7FA}
\definecolor{skillboxframe}{HTML}{4A5A70}
\newtcblisting[auto counter]{skillbox}[2]{%
  enhanced,
  breakable,
  height fixed for=first and middle,
  colback=skillboxbg,
  colframe=skillboxframe,
  colbacktitle=skillboxframe,
  coltitle=white,
  fonttitle=\bfseries\small,
  title={#1},
  label={#2},
  arc=1mm,
  boxrule=0.5pt,
  left=1.5mm,right=1.5mm,top=1mm,bottom=1mm,
  listing only,
  listing options={
    basicstyle=\ttfamily\footnotesize,
    breaklines=true,
    breakatwhitespace=false,
    columns=fullflexible,
    keepspaces=true,
    showstringspaces=false,
    upquote=true,
    frame=none,
    aboveskip=0pt,
    belowskip=0pt,
    extendedchars=true,
    inputencoding=utf8,
    literate=%
      {—}{{\textemdash{}}}1
      {–}{{\textendash{}}}1
      {“}{{\textquotedblleft{}}}1
      {”}{{\textquotedblright{}}}1
      {‘}{{\textquoteleft{}}}1
      {’}{{\textquoteright{}}}1
      {→}{{$\rightarrow$}}1
      {…}{{\ldots{}}}1
      {─}{{-}}1
      {├}{{+}}1
      {└}{{+}}1
      {│}{{|}}1,
  },
}
\crefname{tcb@cnt@skillbox}{Listing}{Listings}
\Crefname{tcb@cnt@skillbox}{Listing}{Listings}

\title{\method{}: Recursive Self-Improvement for Agent Skill Optimization}

\author{%
  Wangcheng Tao \\
  National University of Singapore \\
  \texttt{taowangcheng@u.nus.edu} \\\And
  Zhanglu Yan \\
  National University of Singapore \\
  \texttt{zlyan@nus.edu.sg} \\\AND
  Han Wu\thanks{Corresponding author. Code will be released at \url{https://github.com/taowangcheng/Skill-Infinity}.} \\
  City University of Hong Kong \\
  \texttt{hanwu.cs@my.cityu.edu.hk} \\\And
  Weng-Fai Wong \\
  National University of Singapore \\
  \texttt{wongwf@nus.edu.sg}}

\begin{document}
\maketitle

\begin{abstract}
Recursive self-improvement (RSI), in which a system also improves the way it improves, is a long-standing aspiration on the path to AGI.
Language agents increasingly store procedural knowledge in reusable skills.
Optimizing skills offers a direct way to achieve RSI, because the optimization procedure itself can be encoded as a skill.
Existing skill optimizers, however, follow a hand-designed procedure.
Even methods that evolve a meta-skill confine it to a fixed pipeline and relearn it for each task.
We present \method{}, which makes skill optimization recursively self-improving: a meta agent follows a meta-skill to optimize task skills and optimizes the meta-skill itself using the same procedure.
The optimization is agent-led.
The meta agent diagnoses execution traces, proposes candidates, requests independent reviews, and allocates the search budget.
The meta-skill defines these steps, allowing self-improvement to rewrite them.
This meta-skill is also decoupled from task skills, refined on multiple skill optimization tasks and then reused to optimize task skills across diverse domains.
As in deployment, skills at both levels are standard packages loaded through native skill discovery.
On 83 SkillsBench tasks, \method{} reaches a pass rate of 60.7\%, compared with 56.4\% for human-curated skills and 54.5\%~for~SkillEvolver.
\end{abstract}

\section{Introduction}
\label{sec:intro}

\begin{figure*}[t]
  \centering
  \resizebox{0.97\textwidth}{!}{\begin{tikzpicture}[font=\footnotesize, line cap=round, line join=round]
\definecolor{metaC}{HTML}{2B4C9B}
\definecolor{metaBand}{HTML}{EDF2FD}
\definecolor{metaFill}{HTML}{F1F5FE}
\definecolor{taskC}{HTML}{CF6A18}
\definecolor{taskCx}{HTML}{A65513}
\definecolor{taskBand}{HTML}{FEF3E8}
\definecolor{taskFill}{HTML}{FFF6EC}
\definecolor{priorC}{HTML}{80868F}
\definecolor{priorFill}{HTML}{F1F2F4}
\def\A{3.85}
\def\YS{1.05}
\tikzset{
  declare function={lx(\t)=\A*cos(\t)/(1+sin(\t)^2);
                    ly(\t)=-\YS*\A*sin(\t)*cos(\t)/(1+sin(\t)^2);},
  band/.style={line width=13pt},
  strand/.style={line width=2.2pt},
  tip/.style={-{Stealth[length=8pt,width=7.5pt]}, line width=2.2pt},
  chip/.style={draw, rounded corners=5pt, line width=0.8pt, inner sep=2.2pt, font=\scriptsize},
}
\newcommand{\figfolder}[4]{%
  \begin{scope}[shift={#1}]
    \fill[#3] (-0.40,0.22) -- (-0.40,0.31) -- (-0.17,0.31) -- (-0.10,0.22) -- cycle;
    \node[draw=#3, fill=#4, line width=1pt, rounded corners=1.5pt, minimum width=0.9cm,
          minimum height=0.44cm, inner sep=0pt, text=#3, font=\footnotesize\bfseries] at (0,0) {#2};
  \end{scope}}

\draw[band, metaBand] plot[domain=90:270, samples=181, smooth, variable=\t] ({lx(\t)}, {ly(\t)});
\draw[band, taskBand] plot[domain=-90:90, samples=181, smooth, variable=\t] ({lx(\t)}, {ly(\t)});
\draw[strand, metaC] plot[domain=90:270, samples=181, smooth, variable=\t] ({lx(\t)}, {ly(\t)});
\begin{scope}[scale=1.10]\draw[strand, taskC!40] plot[domain=-90:90, samples=181, smooth, variable=\t] ({lx(\t)}, {ly(\t)});\end{scope}
\begin{scope}[scale=1.05]\draw[strand, taskC!65] plot[domain=-90:90, samples=181, smooth, variable=\t] ({lx(\t)}, {ly(\t)});\end{scope}
\draw[strand, taskC] plot[domain=-90:90, samples=181, smooth, variable=\t] ({lx(\t)}, {ly(\t)});
\foreach \t/\c in {140/metaC, 212/metaC, 320/taskCx, 32/taskCx}
  \draw[tip, \c] ({lx(\t)}, {ly(\t)}) -- ({lx(\t+4)}, {ly(\t+4)});

\figfolder{({lx(113)}, {ly(113)})}{$m_0$}{metaC}{metaFill}
\figfolder{({lx(180)}, {ly(180)})}{$m_1$}{metaC}{metaFill}
\figfolder{({lx(247)}, {ly(247)})}{$m^{\star}$}{metaC}{metaFill}
\foreach \d in {0.16, 0.08} {
  \figfolder{({lx(293)+\d}, {ly(293)+\d})}{}{taskC}{taskFill}
  \figfolder{({lx(0)+\d}, {ly(0)+\d})}{}{taskC}{taskFill}
  \figfolder{({lx(67)+\d}, {ly(67)+\d})}{}{taskC}{taskFill}
}
\figfolder{({lx(293)}, {ly(293)})}{$s_0$}{taskC}{taskFill}
\figfolder{({lx(0)}, {ly(0)})}{$s_1$}{taskC}{taskFill}
\figfolder{({lx(67)}, {ly(67)})}{$s^{\star}$}{taskC}{taskFill}

\node[align=center, text=metaC, font=\large\itshape] at (-2.3, 0) {Meta-skill\\optimizes\\itself\strut};
\node[align=center, text=taskC, font=\large\itshape] at (2.3, 0) {Meta-skill\\optimizes\\task skills\strut};

\node[circle, draw=metaC, fill=metaFill, line width=1.6pt, minimum size=0.86cm] (hub) at (0,0) {};
\fill[metaC] (-0.19,-0.12) rectangle (0.19,0.11);
\fill[metaC] (-0.19,0.11) -- (-0.19,0.17) -- (-0.05,0.17) -- (0.0,0.11) -- cycle;

\node[draw=metaC, fill=metaFill, line width=1pt, rounded corners=4pt, inner sep=5pt,
      anchor=north west, font=\scriptsize] (card) at (4.55, 2.02) {%
  \renewcommand{\arraystretch}{1.05}%
  \begin{tabular}{@{}l@{}}
    {\small\bfseries\ttfamily\color{metaC}meta-skill/}\\[2pt]
    \ttfamily\ \ SKILL.md\\
    \hspace{1.4em}\textit{\color{metaC}\S\ Complete one iteration}\\
    \hspace{1.4em}\textit{\color{metaC}\S\ Analyze the rollout}\\
    \hspace{1.4em}\textit{\color{metaC}\S\ Interpret scores}\\
    {\bfseries\ttfamily\color{metaC}\ references/}\\
    \ttfamily\ \ workspace.md\\
    \ttfamily\ \ evoctl.md\\
    \ttfamily\ \ reviewer-policy.md\\
    {\bfseries\ttfamily\color{metaC}\ scripts/}\\
    \ttfamily\ \ archive.py
  \end{tabular}};
\fill[metaC] ($(card.north west)+(0.18,0)$) -- ++(0,0.13) -- ++(0.95,0) -- ++(0.16,-0.13) -- cycle;
\draw[metaC, line width=0.7pt, dash pattern=on 2.5pt off 2pt] (hub.north) -- (0, 2.45) -| ($(card.north)+(0.5,0.02)$);

\node[font=\small\bfseries] at (-3.45, 1.92) {\method{}:};

\node[anchor=west, text=priorC, font=\scriptsize\bfseries\itshape] at (-7.45, 1.62) {Prior methods:};
\node[chip, draw=priorC, fill=priorFill, text=priorC] (o1) at (-6.75, 1.05) {Optimizer};
\node[chip, draw=taskC, fill=taskFill, text=taskC] (k1) at (-5.12, 1.05) {Skill};
\draw[-{Stealth[length=4pt]}, priorC, line width=0.8pt] (o1) -- (k1);
\node[text=priorC, font=\scriptsize\itshape] at (-5.95, 0.55) {Hand-designed procedure};
\node[chip, draw=priorC, fill=priorFill, text=priorC] (o2) at (-6.75, -0.75) {Meta-skill};
\draw[priorC, line width=0.7pt, dash pattern=on 2pt off 1.6pt, rounded corners=4pt]
  ($(o2.south west)+(-0.16,-0.16)$) rectangle ($(o2.north east)+(0.16,0.42)$);
\draw[-{Stealth[length=3.5pt]}, priorC, line width=0.7pt] ($(o2.north west)+(0.15,0.04)$) to[bend left=60] ($(o2.north east)+(-0.15,0.04)$);
\node[chip, draw=taskC, fill=taskFill, text=taskC] (k2) at (-5.12, -0.75) {Skill};
\draw[-{Stealth[length=4pt]}, priorC, line width=0.8pt] ($(o2.east)+(0.16,0)$) -- (k2);
\node[text=priorC, font=\scriptsize\itshape, align=center] at (-5.95, -1.45) {Evolving in a fixed pipeline\\Bound to one skill};
\end{tikzpicture}}
  \caption{\textbf{Recursively self-improving skill optimization.} \textbf{Left:} most skill optimizers follow a hand-designed procedure. Methods that evolve a meta-skill keep it inside a fixed pipeline and bound to the task skill it evolves with. \textbf{Center:} the ∞ shape shows the two loops of \method{}, which meet at the meta-skill. In the left loop, the meta-skill optimizes itself on a set of skill optimization tasks, improving from $m_0$ to $m^{\star}$. In each stacked loop on the right, the refined meta-skill is reused to optimize one task skill from $s_0$ to $s^{\star}$. \textbf{Right:} the meta-skill at the crossing is a standard skill package loaded through native skill discovery, like any other skill. Each of its parts, from instructions to scripts, falls within the scope of self-improvement.}
  \label{fig:concept}
\end{figure*}
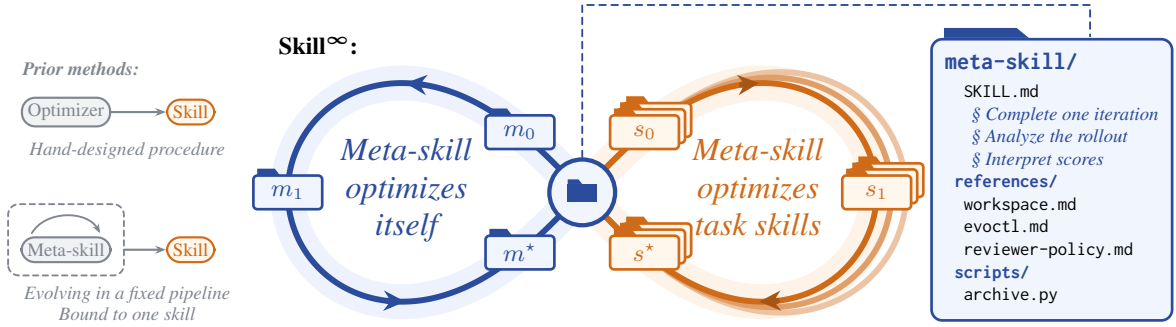

\emph{Recursive self-improvement} (RSI) refers to a system that improves not only its performance on tasks but also the procedure that produces these improvements~\citep{yudkowsky2007, rsisurvey2026}.
Because each gain can then accelerate the next, RSI has figured prominently in visions of artificial general intelligence (AGI), from the intelligence explosion envisioned by \citet{good1966} to Gödel machines~\citep{godelmachine2003}.
More recently, \emph{skills} have emerged as a common format for capturing the reusable procedural knowledge of language agents~\citep{agentskills2025}.
This format can also encode the procedure that optimizes skills.
Skills can thus encode both the target of improvement and the procedure that improves it, making skill optimization a direct and interpretable way to achieve RSI.
Skills also serve as a harness-agnostic medium for RSI, since Claude Code~\citep{claudecode2025}, Codex~\citep{codexcli2025}, and OpenCode~\citep{opencode2025} all load them through the same open specification~\citep{agentskillsspec2025}.
To realize this potential, skill optimization must improve not only skills but also the procedure that optimizes them.

Recent approaches such as SkillOpt~\citep{skillopt2026}, CoEvoSkills~\citep{coevoskills2026}, and SkillEvolver~\citep{skillevolver2026} already optimize skills from execution feedback.
Despite differences in what they edit and how they verify candidates, these methods share one design choice.
The optimizer follows a hand-designed procedure, and what it learns while optimizing one skill does not transfer to the next.
MetaSkill-Evolve~\citep{metaskillevolve2026} goes a step further: each search branch carries a meta-skill that configures the five agents of its pipeline, and the same pipeline refines this meta-skill.
However, the pipeline's roles and wiring remain fixed, and each meta-skill is bound to the task skill it evolves with.
Some methods also optimize skills in a setting that differs from deployment~\citep{skillopt2026, coevoskills2026}.
They inject the skill or prompt the agent to load it, so optimization never tests whether the agent would load the skill on its own.

We believe that an optimizer that improves skills but not its own procedure will eventually stagnate.
RSI targets exactly this limitation.
In this work, we propose \method{}, a framework for recursively self-improving skill optimization.
A meta agent follows a meta-skill to optimize task skills and \emph{applies the same procedure to the meta-skill itself}.
The ``∞'' in \method{} stands for two loops that meet at the meta-skill and feed into each other, as illustrated in \Cref{fig:concept}.
Optimization in \method{} is \emph{agent-led}: in each iteration, the meta agent diagnoses the parent skill from execution traces, proposes and evaluates candidates within the search budget, submits each candidate to an independent reviewer, and selects the next parent.
Self-improvement of the meta-skill refines this procedure, which in turn optimizes task skills more effectively.
We further \emph{decouple} the meta-skill from task skills.
The meta-skill is refined on a set of skill optimization tasks and then reused to optimize diverse task skills.
This decoupling amortizes the cost of self-improvement over task skill optimization.
Skills at both levels are also provided to agents through \emph{native skill discovery}, as in deployment.
We neither inject a skill nor prompt an agent to load one.
Agents see only a skill's name and description and decide whether to load it.
The description is therefore also included in the scope of optimization.

We evaluate \method{} on 83 agentic tasks from SkillsBench~\citep{skillsbench2026}, spanning eight domains, and on GPU kernel optimization from KernelBench~\citep{kernelbench2025}.
In the main comparison, all methods share the Codex CLI~\citep{codexcli2025} as the harness and GPT-6 Luna~\citep{gpt6luna} as the backbone model.
On SkillsBench, \method{} reaches a pass rate of 60.7\%, compared with 56.4\% for human-curated skills and 54.5\% for SkillEvolver, and on KernelBench it reaches a mean speedup of 1.62.
Without the self-improvement of the meta-skill, the pass rate falls to 55.9\%.
\method{} still outperforms human-curated skills when the backbone model or the harness is replaced.
Case studies in Appendix~\ref{app:cases} show that self-improvement changes how the meta agent acts, not only the wording of its instructions.

\section{Related Work}
\label{sec:related}

\paragraph{Recursive Self-Improvement}
The idea of RSI dates back to the argument for an intelligence explosion~\citep{good1966}.
Gödel machines~\citep{godelmachine2003} later formalized it as a program that rewrites its own code once a rewrite is proven useful.
Long a vision of AGI, RSI has since been partly realized in agents built on language models.
STOP~\citep{stop2024} runs a code-improving scaffold on its own code but notes that this is not full RSI, because the language model itself is not altered.
Research on self-evolving agents~\citep{selfevolvingsurvey2025} has introduced systems that rewrite their own logic or code~\citep{godelagent2025, dgm2026, hgm2026} or search over agent designs~\citep{adas2025}.
More recent systems make the meta-level modification procedure itself editable~\citep{hyperagents2026} or compose data, harness, and model updates~\citep{metarsi2026}.
R$^2$ Flow~\citep{r2flow2026} trains a policy that orchestrates a skill library and commits verified edits to it.
These systems implement the optimizer as a program or a trained policy.
\looseness=-1 \method{}, however, encodes the optimizer's procedure as a skill, which makes its recursive self-improvement direct, interpretable, and harness-agnostic.

\paragraph{Agent Skills and Skill Optimization}
Agent skills package reusable procedures as portable directories that agents load on demand~\citep{agentskills2025, agentskillsspec2025, skillsurvey2026}.
This format has precedents in skill libraries, induced workflows, and programmatic web skills~\citep{voyager2024, awm2025, asi2025, skillweaver2025}.
A growing line of work learns skills from experience~\citep{trace2skill2026, skillrl2026} or optimizes them from execution feedback~\citep{evoskill2026, skillopt2026, skillboost2026, skillscoach2026, coevoskills2026, skillhex2026}.
SkillEvolver~\citep{skillevolver2026} writes its optimization procedure as a fixed meta-skill that audits candidate skills in separate agents.
Some methods go further and operate at the meta level, evolving meta-skills that guide skill optimization~\citep{metaskillevolve2026, hisme2026}, multi-agent system design~\citep{skillmas2026}, or context engineering~\citep{mce2026}.
These methods still evolve the meta-skill with a hand-designed procedure.
MetaSkill-Evolve~\citep{metaskillevolve2026}, for instance, fixes the roles and wiring of its five-agent pipeline and binds each meta-skill to the task skill it evolves with.
\method{} instead writes an agent-led optimization procedure into the meta-skill and refines it through self-improvement on several skill optimization tasks.
Decoupled from task skills, the same meta-skill then optimizes diverse task skills.

\paragraph{Prompt and Context Optimization}
Skill optimization builds on prompt optimization.
Early methods select instructions that a language model proposes~\citep{ape2023, opro2024} or evolve populations of prompts~\citep{evoprompt2024, promptbreeder2024}.
Subsequent methods optimize compound systems of model calls~\citep{textgrad2025, dspy2024, gepa2026}, accumulate strategies in evolving contexts~\citep{dynamiccheatsheet2025, ace2026}, or meta-learn system prompts that transfer across tasks~\citep{metaspo2025, sepo2026}.
These methods optimize text that the surrounding system inserts into the model's context.
By contrast, \method{} optimizes skills, which bundle executable scripts and reference files, are loaded only when the agent judges them useful, and can even package the procedure that optimizes them.

\begin{figure*}[t]
  \centering
  \resizebox{\textwidth}{!}{\input{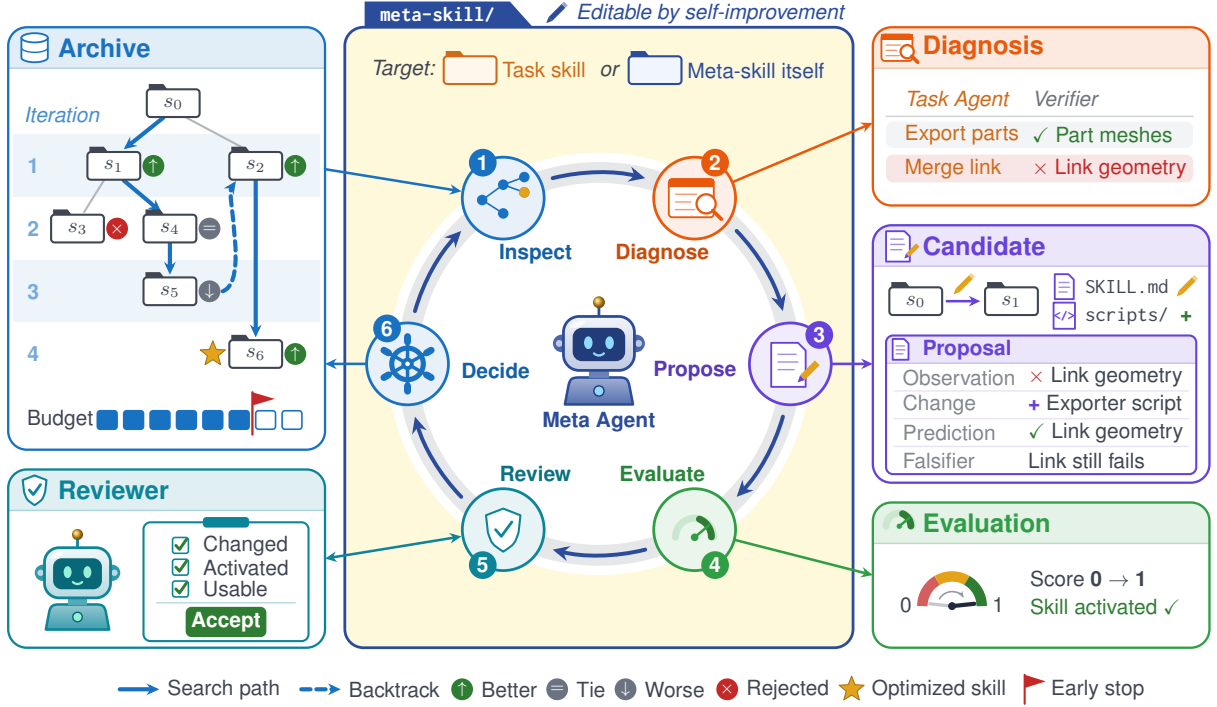}}
  \caption{\textbf{Overview of \method{}.} \textbf{Center:} in each iteration, the meta agent follows six steps defined in the meta-skill, which is editable by self-improvement. The target is a task skill or the meta-skill itself. In the latter case, the meta agent revises the meta-skill while following it. \textbf{Left:} the archive records the optimization history. Drawing on this history, the meta agent steers the search and decides when to stop. An independent reviewer accepts or rejects each candidate. \textbf{Right:} the panels show one iteration on a task that exports the parts of a 3D scene. The meta agent attributes a failed check to one action of the task agent and proposes an exporter script as the fix.}
  \label{fig:overview}
\end{figure*}

\section{Method}
\label{sec:method}

We first formalize skills, native skill discovery, and task skill optimization by a meta-skill.
We then make the meta-skill an optimization target in its own right, so that a single agent-led procedure optimizes task skills and the meta-skill alike.
Finally, we decouple the meta-skill from task skills and extend native skill discovery to both levels.

\subsection{Preliminaries}
\label{sec:method:prelim}

\paragraph{Skill Packages and Native Skill Discovery}
We adopt the Agent Skills specification~\citep{agentskillsspec2025}, an open format through which agents across harnesses share reusable procedural knowledge.
A \emph{skill package} is a directory built around the \texttt{SKILL.md} file.
The front matter of \texttt{SKILL.md} declares the skill's name and description, while the body holds the instructions.
The package may also bundle scripts, reference files, and other resources that the instructions invoke or cite.
With these resources, a skill can provide executable procedures and detailed conventions in addition to textual guidance.
An agent loads skills in stages, following a design principle known as progressive disclosure~\citep{agentskills2025}.
Initially, it sees only the name and description of each installed skill.
For a skill it judges useful for the current task, it then loads the full instructions and opens the bundled resources as needed.
A skill thus takes effect only when the agent \emph{activates} it by loading its instructions.
We term this mechanism \emph{native skill discovery} because it is built into agent harnesses and underlies skill use during deployment.

\paragraph{Tasks and Task Skills}
A \emph{task} $\tau$ consists of an instruction, an environment, and a verifier that returns a scalar reward for each \emph{trial}, one execution by a task agent.
A \emph{task agent} is an agent harness with a backbone model and an installed \emph{task skill} $s$.
The score of $s$ on a set of tasks $T$ is the mean reward over $n$ trials per task, macro-averaged across tasks:
\begin{equation}
\label{eq:score}
S(s; T) = \frac{1}{|T|} \sum_{\tau \in T} \frac{1}{n} \sum_{j=1}^{n} r_j(s, \tau),
\end{equation}
where $r_j(s, \tau)$ is the reward of the $j$-th trial on task $\tau$ with skill $s$.

\paragraph{Task Skill Optimization by a Meta-Skill}
Given a seed task skill $s_0$ and a set of tasks $T$, task skill optimization seeks a skill $s^\star$ that maximizes the score on test tasks.
Our optimizer is a \emph{meta agent} built on the same harness and backbone model as the task agent rather than a stronger one.
Its procedure is itself a skill: the meta agent follows a \emph{meta-skill} $m$, a skill package whose instructions describe how to optimize skills.

\subsection{Recursively Self-Improving Skill~Optimization}
\label{sec:method:rsi}

\paragraph{Skill Optimization Tasks}
We make skill optimization recursively self-improving by optimizing the meta-skill with the same procedure that optimizes task skills.
Just as a task skill is evaluated on tasks, a meta-skill is evaluated on \emph{skill optimization tasks}.
A skill optimization task $q = (s_0^q, T_q, B_q)$ asks the meta agent to optimize a seed task skill $s_0^q$ on tasks $T_q$ within a search budget $B_q$.
Following meta-skill $m$, the meta agent produces an optimized task skill $s^\star_q(m)$ for $q$.
The reward of $m$ on $q$ is the gain of the optimized task skill over the seed:
\begin{equation}
\label{eq:gain}
g(m, q) = S\big(s^\star_q(m); T_q\big) - S\big(s_0^q; T_q\big).
\end{equation}
The \emph{meta score} of $m$ on a set of skill optimization tasks $Q$ is the mean gain:
\begin{equation}
\label{eq:meta-score}
S(m; Q) = \frac{1}{|Q|} \sum_{q \in Q} g(m, q).
\end{equation}
Skill optimization tasks and the meta score are the meta-level counterparts of tasks and the score.
The agent-led procedure of \Cref{sec:method:agentled} therefore applies directly to the meta-skill itself.

\paragraph{The Recursive Loop}
When the meta-skill improves itself, it is both the procedure that the meta agent follows and the target that this procedure optimizes.
The meta agent thus diagnoses and revises this procedure while following it.
The revised meta-skill is then evaluated on skill optimization tasks, where the optimizations it guides reveal whether it surpasses the predecessor.
Such a loop makes self-improvement recursive, since improvements to the meta-skill are self-reinforcing.
Each strengthens the procedure from which further improvements arise, yielding an increasingly capable meta-skill.

\subsection{Agent-Led Optimization}
\label{sec:method:agentled}

\paragraph{Anatomy of an Iteration}
We place the meta agent at the helm of skill optimization, following a meta-skill that is editable by self-improvement.
\Cref{alg:skillrsi} outlines the procedure, and \Cref{fig:overview} depicts one of its iterations.
The meta agent first inspects the optimization history recorded in the archive.
It then diagnoses the parent skill from the execution traces of its rollouts.
Informed by this diagnosis, it creates one or two candidates from the parent, each justified by a proposal.
The meta agent evaluates the candidates and submits each to an independent reviewer.
\looseness=-1 Finally, it decides whether to continue and selects either the next parent or the optimized skill.

\begin{algorithm}[t]
\caption{Agent-led optimization in \method{}}
\label{alg:skillrsi}
\begin{algorithmic}[1]
\Require target skill $s_0$ (a task skill or the meta-skill), meta-skill $m$, tasks $T$, search budget $B$
\Ensure optimized skill $s^\star$
\Statex $\mathcal{A} \gets \{s_0\}$, $p \gets s_0$ \Comment{archive and parent}
\Statex $b \gets B$ \Comment{remaining budget}
\Statex \algorithmicrepeat \Comment{lines 1 to 6: defined in the meta-skill}
\Statex \hspace{\algorithmicindent}Install $m$ for the meta agent, or $p$ if the target is the meta-skill
\State \hspace{\algorithmicindent}Inspect the optimization history in $\mathcal{A}$
\State \hspace{\algorithmicindent}Request rollouts of $p$ and diagnose them
\State \hspace{\algorithmicindent}Create $1 \le k \le \min(2, b)$ candidates from $p$ with proposals, and set $b \gets b - k$
\State \hspace{\algorithmicindent}Evaluate the candidates and add them to $\mathcal{A}$
\State \hspace{\algorithmicindent}Submit each candidate to an independent reviewer
\State \hspace{\algorithmicindent}Decide whether to continue, and select $p$ from $\mathcal{A}$
\Statex \algorithmicuntil\ the meta agent decides to stop or $b = 0$
\Statex \Return $s^\star \gets p$
\end{algorithmic}
\end{algorithm}

\paragraph{Search as Reasoning}
Agent-led optimization entrusts the search to the meta agent's reasoning rather than to a hand-designed heuristic driven by scores.
A score reveals whether a candidate improved but not why, so the meta agent interprets it in light of the diagnoses and reviews recorded in the archive.
The diagnoses can uncover a promising change that the score has yet to reflect, such as a tied candidate that makes progress on the failure it targets.
Much as evolutionary search retains less fit candidates to escape local optima, the meta agent can build on such candidates instead of discarding them.
It may likewise keep the parent when a candidate's gain cannot be attributed to its change, or backtrack to an earlier skill when a line of revisions stalls.
Beyond choosing parents, the meta agent decides which failures to address next, redirecting the search as new evidence emerges.
In sum, search becomes reasoning about the failure modes of each target skill rather than a heuristic applied to all.

\paragraph{Independent Review}
Only candidates that pass an independent review are eligible for selection.
The reviewer operates in an isolated context, shielded from both the meta agent's deliberation and its partiality toward its own changes.
The verdict rests solely on the parent and candidate packages, the proposal, and the supporting evidence.
The reviewer rejects candidates that make no substantive change, that are never activated, or that are unusable as written.
Any gain from such candidates can arise only by chance, and rejecting them prevents selection from mistaking noise for progress.
\looseness=-1 For a meta-skill candidate, the reviewer further verifies that the core responsibilities of a meta-skill, including the review itself, remain intact.

\paragraph{Adaptive Budget Allocation}
The meta agent allocates the search budget itself, deciding how many candidates to create in each iteration and when to stop.
The number of candidates per iteration sets the balance between exploration and exploitation.
Several candidates probe alternative revisions side by side, whereas a single candidate pursues one line of revisions in depth.
Across target skills, a fixed schedule would allocate the same budget regardless of how much headroom remains.
\looseness=-1 The meta agent instead adapts its spending to each target skill, continuing while failures persist and stopping once the diagnosis finds little left to improve.

\paragraph{End-to-End RSI}
The steps of \Cref{alg:skillrsi} are encoded in the meta-skill rather than hard-coded, so the recursive loop of \Cref{sec:method:rsi} can rewrite the procedure itself.
Any part of the package, from the instructions to the bundled scripts, is open to revision.
Freed from any predefined design space, the procedure can evolve beyond the foresight of human designers and realize its full potential.
Self-improvement can remedy a weak step wherever it lies, even in the review and selection of candidates, which a fixed pipeline would leave untouched.
Optimizing task skills reveals the shortcomings of the procedure, and each refinement in turn yields better task skills, rendering the two levels mutually reinforcing.
To keep the mutual reinforcement genuine, scoring lies beyond the reach of self-improvement, so gains can arise only from better optimization, never from reward tampering.

\subsection{Decoupling Meta-Skill from Task Skills}
\label{sec:method:decouple}

\method{} decouples the meta-skill from the task skills it optimizes.
The meta agent first refines a seed meta-skill $m_0$ into $m^\star$ on a set of skill optimization tasks $Q$ drawn from different domains.
Each candidate meta-skill is scored by \Cref{eq:meta-score} on the inner optimizations it guides, one for each $q \in Q$ within the inner budget $B_q$.
The refined meta-skill $m^\star$ then guides the optimization of each target task skill from its seed.

This decoupling separates concerns: the meta-skill captures how to optimize skills, and each task skill how to solve its tasks.
Averaging gains over $Q$, the meta score favors revisions that help across domains over those tailored to one.
As a standard skill package, the refined meta-skill serves all task skills optimized afterwards, amortizing the cost of self-improvement across them.

\subsection{Native Skill Discovery at Both Levels}
\label{sec:method:discovery}

Skills at both levels are installed in the harness's standard skill location and loaded through native skill discovery, as in deployment.
No skill is injected into a prompt, nor is any agent instructed to load one.
Gains measured under these conditions need no injection to materialize in deployment.
For task skills and the meta-skill alike, activation thus hinges on the description, which joins the instructions within the scope of optimization.
Like any skill, the meta-skill is portable across harnesses that support native skill discovery.

\section{Experiments}
\label{sec:experiments}

We evaluate \method{} on SkillsBench~\citep{skillsbench2026} and KernelBench~\citep{kernelbench2025} against baselines with and without skills.
Beyond the main results, we ablate self-improvement and test generality across harnesses and models. We further locate where \method{} helps and measure how performance scales with the search budget and what optimization costs.

\begin{table*}[t]
  \centering
  \small
  \setlength{\tabcolsep}{5pt}
  \begin{tabular}{lccccccccc}
    \toprule
    Method & Overall $\uparrow$ & SWE & Cyber & Science & Finance & Office & Media & Industrial & Math \\
     & (83) & (14) & (7) & (14) & (9) & (14) & (9) & (11) & (5) \\
    \midrule
    No skill                       & 45.5 & 40.0 & 51.4 & 50.0 & 28.9 & 68.6 & 55.6 & 23.6 & 36.0 \\
    Human-curated                  & 56.4 & 47.1 & \textbf{68.6} & \textbf{67.1} & \textbf{37.8} & 70.0 & 68.9 & 41.8 & 40.0 \\
    Self-generated                 & 42.9 & 41.4 & 45.7 & 42.9 & 17.8 & 67.1 & 51.1 & 29.1 & 36.0 \\
    SkillCreator                   & 48.0 & 55.7 & 51.4 & 48.6 & 24.4 & 61.4 & 53.3 & 27.3 & 60.0 \\
    SkillEvolver                   & 54.5 & 60.0 & 42.9 & 52.9 & 28.9 & \textbf{74.3} & 66.7 & 34.5 & 72.0 \\
    \midrule
    \method{} w/o self-improvement & 55.9 & 54.3 & 54.3 & 60.0 & 22.2 & 70.0 & 62.2 & \textbf{52.7} & 68.0 \\
    \method{}                      & \textbf{60.7} & \textbf{65.7} & 51.4 & 57.1 & 26.7 & 70.0 & \textbf{84.4} & 47.3 & \textbf{92.0} \\
    \bottomrule
  \end{tabular}
  \caption{\textbf{Main results on SkillsBench.} Test pass rate (avg@5, \%) overall and per domain, with the number of tasks in parentheses. Domains follow the SkillsBench taxonomy: software engineering (SWE), cybersecurity, natural science, finance and economics, office and white-collar work, media and content production, industrial and physical systems, and mathematics and formal reasoning. All methods use the Codex CLI with GPT-6 Luna.}
  \label{tab:main}
\end{table*}

\subsection{Experimental Setup}
\label{sec:setup}

\paragraph{Benchmarks}
SkillsBench quantifies how much skills improve agent performance across diverse agentic tasks by pairing each task with human-curated skills.
We optimize task skills on the dataset released with SkillEvolver~\citep{skillevolver2026} and test them on its 83 SkillsBench tasks from eight domains.
KernelBench challenges an agent to replace a PyTorch reference model with a faster implementation that computes the same outputs.
Following SkillEvolver, we optimize one task skill shared across three level-3 problems whose reference models are a deep narrow MLP, ShuffleNet, and a GRU.

\paragraph{Baselines}
All baselines use the same harness and backbone model as \method{}.
\textbf{No skill} and \textbf{Human-curated} evaluate the task agent without a skill and with the human-curated skills of SkillsBench, respectively.
\textbf{Self-generated} asks the task agent to write its own skills before using them to solve the task~\citep{skillsbench2026}.
\textbf{SkillCreator} authors a skill with Anthropic's skill-creator~\citep{skillcreator2025}, which CoEvoSkills~\citep{coevoskills2026} adapted to run without human interaction.
\textbf{SkillEvolver}~\citep{skillevolver2026} uses its released meta-skill to author, deploy, and revise a skill over two rounds.

\paragraph{Configurations}
The main experiments use the Codex CLI~\citep{codexcli2025} with GPT-6 Luna~\citep{gpt6luna} at high reasoning effort, and Harbor~\citep{harbor2025} executes the trials.
Seed task skills are instantiated from the template in Appendix~\ref{app:seed}, which identifies the task but carries no task-specific guidance.
We measure the search budget in candidates and cap it at four per task skill.
For self-improvement, the meta agent refines the seed meta-skill in Appendix~\ref{app:meta-skill} on skill optimization tasks drawn from different domains, with the budgets given in Appendix~\ref{app:impl}.
We choose tasks that the seed task skill has not saturated, since a saturated task offers no gain by which a stronger meta-skill could distinguish itself.

\paragraph{Metrics}
As in SkillEvolver, we evaluate performance on SkillsBench by avg@5, the pass rate over five trials per task.
The metric counts a trial as passed when its reward is positive. 
On KernelBench, we report the correctness-weighted speedup, which is the speedup over the PyTorch reference for a correct implementation and zero for an incorrect one.
All reported results are averaged over 5 independent runs.

\subsection{Main Results}
\label{sec:results}

\paragraph{SkillsBench}
Our main comparison asks whether automatically optimized skills can rival those curated by humans.
\Cref{tab:main} reports pass rates on SkillsBench, overall and per domain.
\method{} attains the highest overall pass rate, 60.7\%, surpassing both human-curated skills (56.4\%) and SkillEvolver (54.5\%).
By contrast, SkillCreator (48.0\%) and self-generated skills (42.9\%) remain close to the agent without a skill (45.5\%).
Across domains, \method{} leads in software engineering, media, and mathematics.
From a generic seed, \method{} alone among automatic methods builds task-specific procedural knowledge that serves the agent better than human-curated skills.

\begin{figure}[t]
  \centering
  \begin{tikzpicture}[font=\footnotesize]
\definecolor{oursC}{HTML}{E8590C}
\definecolor{evolverC}{HTML}{1971C2}
\definecolor{noskillC}{HTML}{868E96}
\begin{axis}[
  width=0.96\columnwidth, height=4.5cm,
  set layers, xbar, bar width=4.4pt, reverse legend,
  ytick={0,1,2,3}, yticklabels={Mean, GRU, ShuffleNet, MLP},
  ymin=-0.55, ymax=3.55,
  xmin=0, xmax=2.3, xtick={0,0.5,1,1.5,2},
  xlabel={Correctness-weighted speedup},
  label style={text=black}, tick label style={text=black, /pgf/number format/assume math mode=true},
  axis line style={draw=black!60, line width=0.5pt},
  tick style={draw=black!60}, tick align=outside, major tick length=2pt,
  axis x line*=bottom, axis y line*=left, y axis line style={draw=none}, ytick style={draw=none},
  xmajorgrids, grid style={draw=black!8, line width=0.4pt},
  legend style={at={(0.5,1.02)}, anchor=south, legend columns=3, draw=none, fill=none,
                font=\scriptsize, text=black, /tikz/every even column/.append style={column sep=8pt}},
  legend image code/.code={\fill[#1, rounded corners=0.6pt] (0cm,-0.075cm) rectangle (0.24cm,0.075cm);},
  nodes near coords={\pgfmathprintnumber[assume math mode=true,fixed,fixed zerofill,precision=2]{\pgfplotspointmeta}},
  nodes near coords align={horizontal},
  every node near coord/.append style={font=\scriptsize, inner sep=1.6pt, text=black},
]
  \fill[black!5, on layer=axis background] (axis cs:0,-0.5) rectangle (axis cs:2.3,0.5);
  \addplot[fill=noskillC, draw=none, rounded corners=0.6pt]
    coordinates {(1.02,0) (1.00,1) (1.00,2) (1.05,3)};
  \addplot[fill=evolverC, draw=none, rounded corners=0.6pt]
    coordinates {(1.18,0) (1.02,1) (1.39,2) (1.12,3)};
  \addplot[fill=oursC, draw=none, rounded corners=0.6pt, every node near coord/.append style={font=\scriptsize\bfseries}]
    coordinates {(1.62,0) (1.02,1) (1.97,2) (1.87,3)};
  \legend{No skill, SkillEvolver, \method{}}
  \draw[black!65, dash pattern=on 2pt off 1.6pt, line width=0.65pt] (axis cs:1,-0.55) -- (axis cs:1,3.55);
\end{axis}
\end{tikzpicture}
  \caption{\textbf{Results on KernelBench.} Correctness-weighted speedup over the PyTorch reference, averaged over five trials, on three level-3 problems. The dashed line marks the speed of the reference. Runtimes are measured as in KernelBench on an NVIDIA A100 GPU. All methods use the Codex CLI with GPT-6 Luna.}
  \label{fig:kernel}
\end{figure}
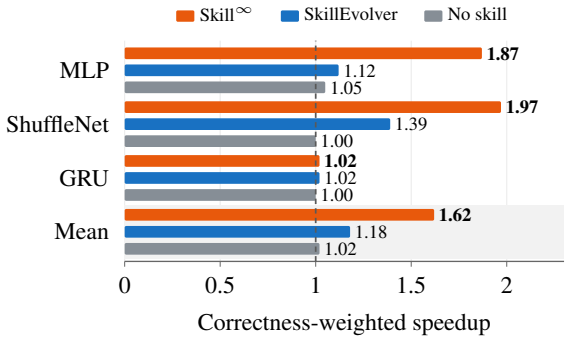

\paragraph{KernelBench}
KernelBench probes the same procedure under an unbounded, continuous reward.
\Cref{fig:kernel} shows the correctness-weighted speedups.
\method{} reaches a mean speedup of 1.62, compared with 1.18 for SkillEvolver and 1.02 without a skill.
On the deep narrow MLP and ShuffleNet, \method{} attains 1.87 and 1.97, against 1.12 and 1.39 for SkillEvolver, respectively.
Beyond whether the agent succeeds, \method{} also improves how well it performs, nearly quadrupling the gain of SkillEvolver over the agent without a skill.

\subsection{Ablation of Self-Improvement}
\label{sec:ablation}

Self-improvement refines the procedure that subsequently optimizes task skills.
To quantify its contribution, we replace the refined meta-skill with its seed and report the result as \method{} w/o self-improvement in \Cref{tab:main}.
Even with the seed meta-skill, \method{} reaches an overall pass rate of 55.9\%, which is higher than those of SkillEvolver and the other automatic baselines.
The refined meta-skill improves the pass rate further to 60.7\%.
Self-improvement helps most in mathematics, media, and software engineering, by 24.0, 22.2, and 11.4 points, respectively.
We attribute these gains to a stricter standard of evidence that self-improvement wrote into the meta-skill.
The refined meta-skill confirms a fix only when the candidate's rollout shows progress on the targeted failure, not when the revised skill merely mentions it.
\looseness=-1 The agent-led procedure alone surpasses the automatic baselines, and by refining it, self-improvement places \method{} above human-curated skills.

\subsection{Generality across Harnesses and Models}
\label{sec:harness}

\begin{figure}[t]
  \centering
  \begin{tikzpicture}[font=\footnotesize]
\definecolor{oursC}{HTML}{E8590C}
\definecolor{curatedC}{HTML}{0C8599}
\begin{axis}[
  width=0.86\columnwidth, height=3.3cm,
  xbar, bar width=6pt, reverse legend,
  ytick={0,1}, yticklabels={{OpenCode\\+ GPT-6 Luna}, {Codex CLI\\+ GPT-5.6 Luna}},
  yticklabel style={align=right},
  ymin=-0.6, ymax=1.6,
  xmin=0, xmax=80, xtick={0,20,40,60,80},
  xlabel={Pass rate (\%)},
  label style={text=black}, tick label style={text=black, /pgf/number format/assume math mode=true},
  axis line style={draw=black!60, line width=0.5pt},
  tick style={draw=black!60}, tick align=outside, major tick length=2pt,
  axis x line*=bottom, axis y line*=left, y axis line style={draw=none}, ytick style={draw=none},
  xmajorgrids, grid style={draw=black!8, line width=0.4pt},
  legend style={at={(0.5,1.02)}, anchor=south, legend columns=2, draw=none, fill=none,
                font=\scriptsize, text=black, /tikz/every even column/.append style={column sep=8pt}},
  legend image code/.code={\fill[#1, rounded corners=0.6pt] (0cm,-0.075cm) rectangle (0.24cm,0.075cm);},
  nodes near coords={\pgfmathprintnumber[assume math mode=true,fixed,fixed zerofill,precision=1]{\pgfplotspointmeta}},
  nodes near coords align={horizontal},
  every node near coord/.append style={font=\scriptsize, inner sep=1.6pt, text=black},
]
  \addplot[fill=curatedC, draw=none, rounded corners=0.6pt]
    coordinates {(53.3,0) (61.2,1)};
  \addplot[fill=oursC, draw=none, rounded corners=0.6pt, every node near coord/.append style={font=\scriptsize\bfseries}]
    coordinates {(56.9,0) (66.5,1)};
  \legend{Human-curated, \method{}}
\end{axis}
\end{tikzpicture}
  \caption{\textbf{Results with another backbone model or harness.} Test pass rate (avg@5, \%) on SkillsBench. Appendix~\ref{app:harness} gives the per-domain results.}
  \label{fig:harness}
\end{figure}
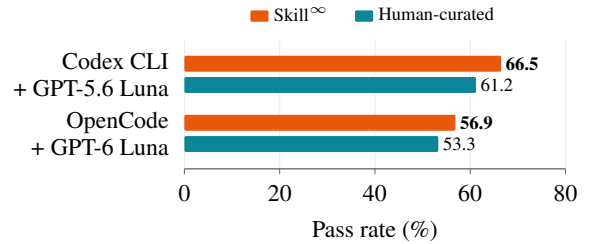

The meta-skill is agnostic to the harness and the backbone model, so \method{} extends to other settings.
We evaluate it in two further settings that replace either the backbone model with GPT-5.6 Luna~\citep{gpt56luna} or the harness with OpenCode~\citep{opencode2025}.
\Cref{fig:harness} compares \method{} with human-curated skills under the same harness and model.
\method{} reaches 66.5\% with GPT-5.6 Luna and 56.9\% with OpenCode, against 61.2\% and 53.3\% for human-curated skills, respectively.
Built on standard skill packages, \method{} keeps its advantage over human-curated skills across harnesses and backbone models.

\subsection{Where \method{} Helps}
\label{sec:utility}

\begin{table}[t]
  \centering
  \small
  \setlength{\tabcolsep}{2.4pt}
  \begin{tabular}{lccccccc}
    \toprule
    Method & A & B1 & B2 & B3 & C1 & C2 & D \\
     & (33) & (8) & (6) & (2) & (13) & (2) & (19) \\
    \midrule
    No skill           & \textbf{95.2} & 37.5 & 36.7 & 60.0 & 0.0 & 0.0 & 0.0 \\
    Human-curated      & 92.1 & 70.0 & 36.7 & 30.0 & \textbf{58.5} & 20.0 & 0.0 \\
    Self-generated     & 85.5 & 50.0 & 30.0 & 40.0 & 1.5 & 20.0 & 1.1 \\
    SkillCreator       & 89.1 & 52.5 & \textbf{43.3} & 10.0 & 6.2 & 30.0 & 10.5 \\
    SkillEvolver       & 91.5 & 65.0 & 33.3 & 20.0 & 15.4 & \textbf{60.0} & 22.1 \\
    \midrule
    \method{}          & 90.9 & \textbf{72.5} & 40.0 & \textbf{70.0} & 29.2 & \textbf{60.0} & \textbf{30.5} \\
    \bottomrule
  \end{tabular}
  \caption{\textbf{Pass rate by skill-utility category.} Test pass rate (avg@5, \%) on SkillsBench, with the number of tasks in parentheses. Appendix~\ref{app:data} defines the categories.}
  \label{tab:category}
\end{table}

An overall pass rate does not show which tasks benefit from \method{}.
Following SkillEvolver~\citep{skillevolver2026}, we group tasks by skill utility as defined in Appendix~\ref{app:data}.
Category A holds tasks that the agent solves without a skill.
B1, B2, and B3 hold partly solved tasks on which human-curated skills improve, preserve, or degrade the pass rate, respectively.
C1 and C2 hold tasks that only human-curated skills solve, at pass rates of at least 40\% and below 40\%, respectively.
D holds tasks that neither the agent alone nor human-curated skills solve.
\Cref{tab:category} reports the pass rate per category.
The improvements of \method{} concentrate on tasks beyond the agent's reach.
On tasks that neither the agent nor human-curated skills solve (D), \method{} reaches 30.5\%, against 22.1\% for SkillEvolver.
On tasks that only human-curated skills solve at a pass rate of at least 40\% (C1), \method{} reaches 29.2\%, against 15.4\% for SkillEvolver.
\method{} provides the procedural knowledge that the agent lacks, even on tasks that human-curated skills fail to solve.

\subsection{Budget Scaling and Amortized Cost}
\label{sec:budget}

\begin{figure}[t]
  \centering
\begin{tikzpicture}[font=\footnotesize]
\definecolor{oursC}{HTML}{E8590C}
\definecolor{noskillC}{HTML}{868E96}
\begin{axis}[
  width=\columnwidth, height=4.3cm,
  set layers,
  xmode=log, log basis x=2, xmin=0.8, xmax=10,
  xtick={1,2,4,8}, xticklabels={1,2,4,8},
  ymin=0, ymax=100, ytick={0,20,40,60,80,100},
  xlabel={Search budget (candidates)},
  ylabel={Pass rate (\%)},
  label style={text=black}, tick label style={text=black, /pgf/number format/assume math mode=true},
  axis line style={draw=black!60, line width=0.5pt},
  tick style={draw=black!60}, tick align=outside, major tick length=2pt,
  axis x line*=bottom, axis y line*=left,
  ymajorgrids, grid style={draw=black!8, line width=0.4pt},
]
  \addplot[name path=base, draw=noskillC, dashed, line width=0.9pt, domain=0.8:10, samples=2] {12.5};
  \node[anchor=north east, text=black, inner sep=1.5pt] at (axis cs:10,12.5) {No skill};
  \addplot[name path=curve, draw=oursC, line width=1.6pt, line join=round, line cap=round, mark=*, mark size=2.4pt,
           mark options={fill=white, draw=oursC, line width=1.2pt},
           nodes near coords={\ifnum\coordindex=2 default\fi},
           every node near coord/.append style={anchor=south, yshift=2.5pt, text=oursC!85!black, font=\scriptsize\itshape}]
    coordinates {(1,47.5) (2,70.0) (4,77.5) (8,82.5)};
  \addplot[fill=oursC!16, draw=none, on layer=axis background]
    fill between[of=curve and base, soft clip={domain=1:8}];
\end{axis}
\end{tikzpicture}
  \caption{\textbf{Pass rate under different search budgets.} Test pass rate (avg@5, \%) on a subset of SkillsBench tasks. The solid line shows the optimized task skills, and the dashed line the agent without a skill.}
  \label{fig:budget}
\end{figure}
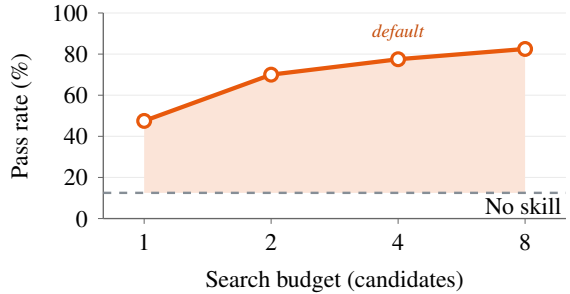

\paragraph{Scaling the Search Budget}
We ask whether a larger budget yields further gains, given that the meta agent allocates the budget itself.
We run this analysis on a subset of tasks, where the default budget substantially improves the seed task skill.
We then optimize their task skills with budgets of one, two, and eight candidates under otherwise identical settings.
As shown in \Cref{fig:budget}, the pass rate rises with each doubling of the budget, from 47.5\% with one candidate to 82.5\% with eight.
Even with a single candidate, the optimized task skills far exceed the agent without a skill (12.5\%).
The meta agent allocates each budget itself, so these gains require no retuning of the procedure.
\looseness=-1 Agent-led optimization scales with the search budget.

\paragraph{Cost}
We measure cost by the tokens that all agents consume during optimization, as itemized in Appendix~\ref{app:cost}.
Optimizing a task skill with the refined meta-skill requires 8.7M input and 66k output tokens on average, close to 8.1M and 76k for SkillEvolver.
The cost of self-improvement is amortized across all task skills optimized afterwards.
Self-improvement consumes about as many tokens as eleven task skill optimizations.
Spread over the 83 task skills, this cost adds less than 15\% to each optimization, a share that shrinks with each further reuse of the meta-skill.
By reusing its refined meta-skill, \method{} surpasses SkillEvolver at a comparable cost.

\section{Conclusion}
\label{sec:conclusion}

Skill optimizers have so far followed hand-designed procedures.
Even methods that evolve a meta-skill fix the pipeline around it and relearn it for each task.
We presented \method{}, which makes skill optimization recursively self-improving.
A meta agent follows a meta-skill to optimize task skills, and the same agent-led procedure optimizes the meta-skill itself.
The meta-skill, decoupled from task skills, is refined on skill optimization tasks and then reused to optimize task skills across domains.
At both levels, agents load their skills through native skill discovery, as in deployment.
On SkillsBench, \method{} outperforms human-curated skills and SkillEvolver, and it improves kernel speedups on KernelBench.
Encoding the optimizer's procedure as a skill makes recursive self-improvement direct, interpretable, and harness-agnostic.
We hope that this view invites further work on agents that improve the way they improve.

\section*{Limitations}

\method{} makes skill optimization recursively self-improving while the backbone model stays fixed.
The resulting gains are therefore bounded by the expressiveness of skill packages.

Both levels of optimization rely on automatic verifiers, since the meta score is computed from task rewards.
\method{} therefore applies to domains in which such rewards can be computed reliably.

Finally, agent-led optimization trades enforcement for flexibility.
The meta agent follows its procedure rather than executing it as code.
The effect of the meta-skill thus depends on how reliably the backbone model follows long instructions.
No tool enforces this adherence, which rests instead on the meta-skill and on independent review.

\section*{Ethical Considerations}

Optimized skills bundle instructions and executable scripts that agents follow with tool access, and an agent may activate them without user approval.
We therefore recommend reviewing what these skills do and when they are activated before deploying them.
Some SkillsBench tasks concern cybersecurity and are used only in the form the benchmark provides.
SkillsBench is released under the Apache 2.0 license, and KernelBench and the dataset of SkillEvolver under the MIT license.
We use these artifacts for research evaluation, consistent with their intended use.

\bibliography{references}

\appendix
\section{Skill-Utility Categories}
\label{app:data}

Following SkillEvolver, we assign tasks to skill-utility categories by the pass rates of No skill and Human-curated in \Cref{tab:main}.
Category A contains tasks on which No skill reaches a pass rate of at least 80\%.
Categories B1, B2, and B3 contain tasks on which the pass rate of No skill lies strictly between 0\% and 80\%.
On these tasks, Human-curated achieves a higher, equal, or lower pass rate, respectively.
Categories C1 and C2 contain tasks that No skill never passes.
On these tasks, Human-curated reaches at least 40\% or remains strictly between 0\% and 40\%, respectively.
Category D contains tasks that neither method passes.

\section{Implementation Details}
\label{app:impl}

The inner budget is two candidates per skill optimization task, and self-improvement may create at most four meta-skill candidates.
In each iteration, the meta agent receives the request below.

\begin{skillbox}{Request to the meta agent}{lst:request}
Optimize the Task Skill in this workspace.
Workspace: <absolute path>
Complete optimization iteration <id>, write its decision, and then return.
\end{skillbox}

\section{Case Studies}
\label{app:cases}

\paragraph{The Meta-Skill Rewrites Its Selection Rule}
During self-improvement, the meta agent diagnosed the seed meta-skill from the inner optimizations it had guided.
For the 3D scene task, the inner optimization had accepted two consecutive candidates while the verifier still reported the same link-geometry failure.
It had also selected the second candidate, which only tied the parent.
The meta agent located this weakness in the inspection and selection of evaluated candidates rather than in diagnosis.
The candidate meta-skill it wrote requires checking the rollout of each tied or worse candidate for the targeted failure.
A tied candidate can then be selected only when that rollout shows progress.
The candidate improved the meta score from 0.55 to 0.83.

In the next iteration, the meta agent followed this candidate and proposed a further change that tracks intermediate scores and verifier failures across candidates.
The new candidate tied its parent.
Applying the rule it had just written, the meta agent selected the tied candidate only after checking its inner optimization on the 3D scene task.
There, the targeted failure had been addressed, and the task skill had reached a perfect score.
Self-improvement then stopped after two further candidates and retained this candidate as the refined meta-skill.
Across the 83 task skill optimizations, the refined meta-skill selected 54\% of tied candidates, against 83\% for the seed meta-skill.
Self-improvement changes the behavior of the meta agent, not merely the text of its instructions.

\paragraph{A Task Skill Gains a Script}
For a SkillsBench task that exports the parts of a Three.js scene as OBJ files, the seed task skill fails all test trials.
In the first iteration, a rollout of the seed showed that each individual mesh passed while a merged link file failed the verifier's geometry check.
The meta agent attributed this failure to group ownership and coordinate transforms.
It added an exporter script, wrote the ownership and transform rules into \texttt{SKILL.md}, and rewrote the description to specify the files the skill produces.
Its proposal states a falsifying outcome in terms of the verifier's check.
The reviewer confirmed that the script exists and matches the documented interface before accepting the candidate.
The optimized skill passes all five test trials.
The boxes below show the optimized task skill and its proposal.

\begin{skillbox}{Optimized task skill (\texttt{SKILL.md}, packaged with \texttt{scripts/export\_scene.mjs})}{lst:threejs-final}
---
name: threejs-structure-parser
description: Export named Three.js group parts and transformed OBJ mesh and link files. Load before inspecting files or running commands.
---

# Task Execution

Read the task instruction and inspect every provided input before changing files. Treat the scene hierarchy and verifier as the source of truth for what counts as a part, how nested meshes are assigned, and which coordinate space exports must use.

For a scene module that exports `createScene()`, use the bundled exporter:

```bash
node /root/.agents/skills/threejs-structure-parser/scripts/export_scene.mjs /root/data/object.js /root/output
```

Replace the paths with the task's actual input module and output directory. The script requires `three` to be resolvable from the input module's location. For other module formats or output formats, follow the same rules below in an appropriate exporter.

## Export rules

- A mesh belongs to its nearest named ancestor group. Do not assign nested-group meshes to every ancestor. Export each nonempty group as a part directory and a combined link file.
- Apply each mesh's full world matrix before writing either an individual mesh OBJ or a merged link OBJ. Group-local vertices omit parent transforms and will misplace parts in world space.
- Write one individually named OBJ for every mesh in its assigned part directory. Merge the transformed geometries assigned to that group for its link OBJ; preserve all component vertices.
- Keep names and directory structure required by the instruction. Detect duplicate group or mesh output names instead of silently overwriting files.

Before finishing, inspect the exporter output and validate it against the task's supplied tests or verifier. Check that all expected group directories and mesh filenames exist, each OBJ contains vertices and faces, and link geometry matches the required coordinate space and membership. Read verifier failures and fix their underlying cause before reporting completion.
\end{skillbox}

\begin{skillbox}{Proposal written by the meta agent for the optimized skill}{lst:threejs-proposal}
{
  "observation": "Training activation was confirmed and all 60 individual part OBJs passed, but link validation failed: slewing_unit.obj had Chamfer distance 103.0 against a 0.0002 threshold. The rollout exporter selected only direct children and kept group-local transforms; the supplied ground-truth generator assigns each mesh to its nearest named ancestor group and applies mesh.matrixWorld before merging.",
  "change": "Added a reusable Three.js OBJ export script and explicit skill guidance for nearest-named-group ownership, full world-matrix transforms on individual and merged geometry, duplicate-name checks, and verifier-guided validation.",
  "predicted_effect": "The exporter will retain the passing per-mesh output while making link OBJ membership and coordinates match the verifier's expected world-space aggregates, eliminating the observed link Chamfer failure.",
  "falsifying_outcome": "Validation still reports a link OBJ Chamfer distance at or above 0.0002, especially for a group with a translated or nested parent, or the exporter fails to produce the expected mesh and link filenames."
}
\end{skillbox}

\section{Results with Another Model or Harness}
\label{app:harness}

\Cref{tab:harness-domains} reports the results of \Cref{sec:harness} by domain.

\begin{table*}[t]
  \centering
  \small
  \setlength{\tabcolsep}{5pt}
  \begin{tabular}{lccccccccc}
    \toprule
    Method & Overall $\uparrow$ & SWE & Cyber & Science & Finance & Office & Media & Industrial & Math \\
     & (83) & (14) & (7) & (14) & (9) & (14) & (9) & (11) & (5) \\
    \midrule
    \multicolumn{10}{l}{\textit{Codex CLI with GPT-5.6 Luna}} \\
    Human-curated                  & 61.2 & 40.0 & \textbf{74.3} & \textbf{71.4} & 37.8 & 64.3 & \textbf{77.8} & \textbf{61.8} & 76.0 \\
    \method{}                      & \textbf{66.5} & \textbf{54.3} & 62.9 & 68.6 & \textbf{40.0} & \textbf{87.1} & 75.6 & 60.0 & \textbf{88.0} \\
    \midrule
    \multicolumn{10}{l}{\textit{OpenCode with GPT-6 Luna}} \\
    Human-curated                  & 53.3 & 52.9 & 65.7 & \textbf{58.6} & \textbf{28.9} & 62.9 & \textbf{64.4} & 38.2 & 52.0 \\
    \method{}                      & \textbf{56.9} & \textbf{67.1} & \textbf{68.6} & 57.1 & 22.2 & \textbf{72.9} & 55.6 & \textbf{41.8} & \textbf{64.0} \\
    \bottomrule
  \end{tabular}
  \caption{\textbf{Per-domain results with another backbone model or harness.} Test pass rate (avg@5, \%) on SkillsBench, with the number of tasks in parentheses. Domains follow \Cref{tab:main}.}
  \label{tab:harness-domains}
\end{table*}

\section{Optimization Cost}
\label{app:cost}

\Cref{tab:cost} lists the tokens consumed while optimizing a task skill, together with those of self-improvement.
Neither Self-generated nor Human-curated appears in the table, since the former writes its skills within the test trials and the latter needs no optimization.

\begin{table}[ht]
  \centering
  \small
  \setlength{\tabcolsep}{4.5pt}
  \begin{tabular}{lccc}
    \toprule
    Method & Input (M) & Cached (M) & Output (k) \\
    \midrule
    \method{} & 8.74 & 8.23 & 66.3 \\
    SkillEvolver & 8.06 & 7.64 & 75.8 \\
    SkillCreator & 2.86 & 2.69 & 30.6 \\
    \midrule
    \multicolumn{4}{l}{\textit{Self-improvement}} \\
    \quad Total & 97.65 & 92.00 & 778.6 \\
    \quad Per task skill & 1.18 & 1.11 & 9.4 \\
    \bottomrule
  \end{tabular}
  \caption{\textbf{Tokens consumed during optimization.} Means per task skill over the 83 SkillsBench tasks with the Codex CLI and GPT-6 Luna. Token counts cover all agents involved in optimization and exclude the test trials. Input tokens include cached ones. Self-improvement is given in total and spread over the 83 task skills.}
  \label{tab:cost}
\end{table}

\section{Seed Task Skills}
\label{app:seed}

All SkillsBench seed task skills are instantiated from a single template.
Its description identifies the task, and its body prescribes only generic inspection, execution, and validation, with no knowledge specific to the task.
The KernelBench task skill starts from a similar seed whose description refers to KernelBench tasks and whose body provides generic advice on profiling, correctness checks, and timing.

\section{Seed Meta-Skill}
\label{app:meta-skill}

The seed meta-skill is the starting point of self-improvement, and \method{} w/o self-improvement uses it to optimize all task skills.
Its package contains \texttt{SKILL.md}, three reference files, and a Python script.
\texttt{SKILL.md} defines the iteration of \Cref{alg:skillrsi}, the diagnostic procedure for both kinds of targets, and the scores.
The reference files define the workspace and the decision record, the interface of the rollout and evaluation tools, and the review criteria.
The script assembles the archive from the files in the workspace and initializes each candidate as a copy of its parent.
\texttt{SKILL.md} and the reference files on the workspace and the tools are reproduced verbatim below.
The front matter retains the package's internal name.
The names \texttt{Meta Agent}, \texttt{Task Agent}, and \texttt{Reviewer Subagent} in the files correspond to the meta agent, task agent, and reviewer in the paper, respectively.

\begin{skillbox}{Seed meta-skill: \texttt{SKILL.md}}{lst:seed-skill}
---
name: skill-self-evolver
description: Optimize a Task Skill or this Meta Skill from rollout reports and evaluation results through candidate generation, scoring, review, and selection.
---

# Optimize skills

Act as the Meta Agent in the supplied workspace.

Read this skill's instruction references before applying them.

The [workspace guide](references/workspace.md) defines candidates, the archive, the candidate budget, and decisions. The [evoctl interface](references/evoctl.md) defines evaluation commands and their results.

## Complete one iteration

1. **Inspect the archive.** Inspect the archive for the optimization target, parent, active iteration, prior outcomes, and remaining candidate budget.

2. **Diagnose the parent.** Obtain its rollout with `evoctl rollout`, and also the rollouts of its children from earlier iterations that scored no higher than it. Complete the [diagnostic procedure](#analyze-the-rollout), then form falsifiable improvement hypotheses.

3. **Create candidates.** Create one or two candidates in the active iteration within the remaining candidate budget, each containing a complete copy of the parent's `skill/` directory. Use the rollout diagnosis to improve the skill's instructions and resources. Keep the skill's name unchanged and its resource links valid. Write `proposal.json` in each candidate directory with the supporting observation, change, predicted effect, and falsifying outcome. A Meta Skill candidate preserves the [required responsibilities](references/reviewer-policy.md#meta-skill-responsibilities).

4. **Evaluate candidates.** Evaluate ready candidates in parallel with `evoctl evaluate`. Obtain and read each completed result.

5. **Review evaluated candidates.** For each scored candidate, start a fresh Reviewer Subagent without inheriting your conversation. Provide absolute paths to the parent directory, the candidate directory, the candidate's `evaluation.json`, the parent's `rollout.md`, and the currently installed [reviewer policy](references/reviewer-policy.md). Run ready reviews alongside independent work. Wait for each reviewer to finish and read the candidate's `review.json`; use the same reviewer for corrections to that review.

6. **Select and return.** After every submitted evaluation and the review of every scored candidate finish, refresh the archive. Compare the validation scores of the parent and reviewed candidates, considering reviews, history, and remaining candidate budget, then select whichever `seed` or `accepted` node you judge best. Select a candidate that only ties its parent when its change addresses a failure or requirement identified in the diagnosis. Write the active iteration's [decision.json](references/workspace.md#record-the-decision), then return.

## Analyze the rollout

Analyze the parent's rollout to decide how to improve its skill. Treat the supplied reports, complete task packages including tests and reference solutions, and exported artifacts as read-only materials.

Start with the rollout overview and follow relevant trajectory links. Use search and contextual reading to verify the actions and tool results behind an outcome, including subsequent recovery. Expand your reading when evidence is incomplete or conflicting.

### Task Skill

1. **Understand the tasks and rewards.** Identify the skill's supplied training tasks in the report. Read their instructions and tests/verifier source, including helpers that define the reward conditions. Inspect inputs and reference solutions as needed to understand the required behavior.

2. **Analyze the execution.** Choose trials covering the relevant success and failure outcomes. Match the Task Agent's actions, tool results, recovery attempts, and final outputs to the task requirements and verifier checks, and note avoidable detours.

3. **Identify skill improvements.** Compare the Task Skill's guidance with the Task Agent's actual behavior. Identify missing, unclear, misleading, or unnecessary guidance, separating skill issues from execution, environment, and verifier issues. When the Task Agent rewrites or gets wrong a procedure that code can perform, provide that procedure as a script; when it needs detailed conventions or data, provide them in a reference file linked from `SKILL.md`. Propose reusable changes supported by the recorded evidence, preserving guidance that worked.

### Meta Skill

1. **Locate optimization strengths and weaknesses.** Compare every supplied Task Skill's initial and final validation scores, gains, and the Meta Skill's training score. Select optimization trajectories that explain improvements, regressions, or missed opportunities.

2. **Analyze the optimization process.** Trace how the Meta Agent and Reviewer Subagent used evidence, modified skills, and responded to feedback through their messages, actions, and tool results. Consult the Task Skill versions, available Task Skill rollouts, and node and iteration records linked from `rollout.md` as needed to verify their diagnoses and changes.

3. **Identify Meta Skill improvements.** Relate the current Meta Skill's instructions to the observed diagnosis, candidate generation, review, and selection. Determine which guidance would help the Meta Agent produce better Task Skills across the supplied tasks.

## Interpret scores

The score, `S(skill; evaluation_set)`, aggregates performance on a fixed training or validation set. A gain is the difference between two scores under the same evaluation conditions.

For a **Task Skill**, the evaluation set contains tasks. The verifier produces a reward for each trial:

```text
task_score(s, t) = mean of trial rewards for task t
S(s; T) = mean of task_score(s, t) over the supplied tasks t in T
```

A `timeout` trial is scored on the artifacts present when the Task Agent reached its deadline; trials without a verifier reward, including `verifier_error`, count as zero when averaging trial rewards.

For a **Meta Skill**, the evaluation set contains skill optimization tasks. Each optimization task provides an initial Task Skill with its training and validation sets, and fixed optimization conditions. The Meta Agent uses the Meta Skill and its reviewer policy to optimize that Task Skill. Measure the improvement in its validation score:

```text
gain(m, q) = S(final Task Skill produced by m; V_q) - S(initial Task Skill; V_q)
S(m; Q) = mean of gain(m, q) over the supplied skill optimization tasks q in Q
```

Here `V_q` is the validation set for the Task Skill in optimization task `q`.
\end{skillbox}

\begin{skillbox}{Seed meta-skill: \texttt{references/workspace.md}}{lst:seed-workspace}
# Optimization workspace

`WORKSPACE` is the path supplied in the optimization request. `context.json` supplies `optimization_target` (`task_skill` or `meta_skill`) and this run's candidate budget (`budget.max_candidates_per_run`). Each node holds one complete skill version.

`META_SKILL_DIR` is the directory containing the installed `SKILL.md` you are running. Use that location for this skill's scripts and references.

```text
WORKSPACE/
├── context.json
├── tasks/<task-skill-id>/<task-id>/  # Training task packages
├── seed/
│   ├── node.json
│   ├── skill/<skill-name>/
│   ├── evaluation.json
│   ├── rollout.md                # Created by a rollout request
│   └── rollout/
└── iterations/000/
    ├── candidates/i000-c1/
    │   ├── node.json
    │   ├── skill/<skill-name>/
    │   ├── proposal.json
    │   ├── evaluation.json
    │   ├── review.json
    │   ├── rollout.md           # Created by a rollout request
    │   └── rollout/
    └── decision.json
```

The archive is derived from these files. Use the currently installed [archive helper](../scripts/archive.py), or read and organize the files directly.

## Task Skill rollout

`rollout.md` summarizes training results and links to each task trial's trajectory and supporting materials:

```text
<node>/rollout/<task-id>/<trial-index>/
├── trajectory.md               # Task Agent and subagent executions
├── verifier/                   # Recorded verifier results and diagnostics
├── artifacts/                  # Collected task outputs
└── media/                      # Recorded images
```

## Meta Skill rollout

`rollout.md` summarizes training results and links to the optimization trajectory and supporting materials for each Task Skill being optimized:

```text
<node>/rollout/<task-skill-id>/
├── trajectory.md               # Meta Agent and Reviewer Subagent executions
├── nodes/<node-id>/             # Task Skill, node records, and available rollout reports
├── iterations/                 # Per-iteration decisions and other recorded files
└── media/                      # Recorded images
```

## Inspect the archive

```bash
python META_SKILL_DIR/scripts/archive.py view --workspace WORKSPACE
```

The archive view includes `context` from `context.json`, nodes, scores, review status, generated-candidate counts, previous decisions, the active parent, `candidate_root`, `decision_path`, and `remaining_candidates`. Creating a candidate consumes one candidate unit, including an unevaluated draft; keep it in the archive.

## Create a candidate

Choose a new directory under the active `candidate_root`. Name it `iNNN-cK`, where `NNN` is the iteration number and `K` is a positive candidate number. Use the active parent as the source for this iteration.

```bash
python META_SKILL_DIR/scripts/archive.py create --workspace WORKSPACE --parent-dir PARENT_DIR --candidate-dir CANDIDATE_DIR
```

The helper copies `PARENT_DIR/skill/` to `CANDIDATE_DIR/skill/`, writes `CANDIDATE_DIR/node.json`, and returns the candidate's paths. Edit the copied skill and write `CANDIDATE_DIR/proposal.json`. Node metadata has this form, with `name` identifying the directory under `skill/`:

```json
{
  "node_id": "i000-c1",
  "parent_id": "seed",
  "iteration_id": "i000",
  "name": "skill-name"
}
```

A scored and reviewed candidate is `accepted` or `rejected` according to its review; unresolved candidates are `proposed`. The archive exposes each scored node's `validation_score` from its `evaluation.json`.

## Record the decision

Write one `decision.json` in the active iteration:

```json
{
  "action": "continue",
  "selected_node_id": "i000-c1",
  "reason": "Evidence-grounded selection rationale and next optimization direction."
}
```

Write exactly these three keys. Use `continue` when the run continues: `selected_node_id` becomes the next iteration's parent. Use `stop` when the run ends, because the candidate budget is exhausted or you choose to finish: `selected_node_id` is the run's final skill. If the candidate budget is later raised, a new iteration continues from that node. Ground the selection in the evidence in `reason`: for `continue`, add the next optimization direction; for `stop`, add why the search ends here.
\end{skillbox}

\begin{skillbox}{Seed meta-skill: \texttt{references/evoctl.md}}{lst:seed-tools}
# evoctl evaluation tools

Successful commands return JSON. A correctable input error returns JSON with an `error` message string and a nonzero exit status. Keep a node's skill unchanged once submitted to either command; further edits belong in a new candidate.

## Obtain a rollout

```bash
evoctl rollout --workspace WORKSPACE --node-dir NODE_DIR
```

`NODE_DIR` identifies the skill node to run on the supplied training set. The response contains `optimization_target` (as defined in the [workspace guide](workspace.md)), `node_id`, and `rollout_path`. `rollout_path` points to `NODE_DIR/rollout.md`, which provides the training score, skill activation, usage, elapsed time, and links to trajectories and supporting materials.

Skill activation is reported as `Yes`, `No`, or `Unknown`.

## Evaluate a candidate

```bash
evoctl evaluate --workspace WORKSPACE --candidate-dir CANDIDATE_DIR
```

`CANDIDATE_DIR` identifies a [candidate node](workspace.md#create-a-candidate).

The command evaluates the candidate on the supplied validation set and returns JSON containing `node_id`, `validation_score`, and `skill_activated`. The result is saved to `CANDIDATE_DIR/evaluation.json`.

`skill_activated` is `true` when activation is confirmed across the evaluated executions, `false` when any execution lacks activation, and `null` when unknown.
\end{skillbox}

\end{document}